\documentclass[11pt, reprint,
    superscriptaddress,
    longbibliography,
    amsmath,
    amssymb,
    aps,
    floatfix
]{revtex4-2}

\usepackage{graphicx}
\usepackage{dcolumn}
\usepackage{bm}
\usepackage[utf8]{inputenc}
\usepackage{acronym}
\usepackage{graphicx}
\usepackage{amsmath,amsfonts,amssymb}
\usepackage{braket}
\usepackage{tikz}
\usepackage[table]{xcolor}
\usetikzlibrary{quantikz2}
\usepackage{acronym}
\usepackage{xspace}
\usepackage{siunitx}
\usepackage{booktabs}
\usepackage[hidelinks]{hyperref}
\usepackage{nicefrac}       
\usepackage{microtype}      
\usepackage{mathtools}
\usepackage{multirow}
\usepackage{adjustbox}
\usepackage{listings}
\usepackage{siunitx}
\usepackage{subfigure}

\begin{document}

\title{Towards Tensor Network Models for Low-Latency Jet Tagging on FPGAs}

\author{Alberto Coppi$^\dagger$}
\thanks{Contact author: \href{mailto:alberto.coppi@phd.unipd.it}{alberto.coppi@phd.unipd.it}}
\affiliation{Dipartimento di Fisica e Astronomia "Galileo Galilei", University of Padua, 35131 Padova, Italy}
\affiliation{Istituto Nazionale di Fisica Nucleare (INFN), 35131 Padova, Italy}

\author{Ema Puljak}
\thanks{These authors contributed equally to this work.}
\affiliation{Universitat Autònoma de Barcelona, 08193 Bellaterra (Barcelona), Spain}

\author{Lorenzo Borella}
\thanks{These authors contributed equally to this work.}
\affiliation{Dipartimento di Fisica e Astronomia "Galileo Galilei", University of Padua, 35131 Padova, Italy}
\affiliation{Istituto Nazionale di Fisica Nucleare (INFN), 35131 Padova, Italy}

\author{Daniel Jaschke}
\affiliation{PlanQC GmbH, Lichtenbergstr. 8, 85748 Garching, Germany}
\affiliation{Institute for Complex Quantum Systems, Ulm University, 89069 Ulm, Germany}
\affiliation{Dipartimento di Fisica e Astronomia "Galileo Galilei", University of Padua, 35131 Padova, Italy}
\affiliation{Istituto Nazionale di Fisica Nucleare (INFN), 35131 Padova, Italy}

\author{Enrique Rico}
\affiliation{European Organization for Nuclear Research (CERN), CH-1211 Geneva, Switzerland}

\author{Maurizio Pierini}
\affiliation{European Organization for Nuclear Research (CERN), CH-1211 Geneva, Switzerland}

\author{Jacopo Pazzini}
\affiliation{Dipartimento di Fisica e Astronomia "Galileo Galilei", University of Padua, 35131 Padova, Italy}
\affiliation{Istituto Nazionale di Fisica Nucleare (INFN), 35131 Padova, Italy}
\affiliation{Dipartimento di Ingegneria dell'Informazione, Università di Padova, Italy}
\affiliation{Dipartimento di Ingegneria Industriale, Università di Padova, Italy}

\author{Andrea Triossi}
\affiliation{Dipartimento di Fisica e Astronomia "Galileo Galilei", University of Padua, 35131 Padova, Italy}
\affiliation{Istituto Nazionale di Fisica Nucleare (INFN), 35131 Padova, Italy}

\author{Simone Montangero}
\affiliation{Dipartimento di Fisica e Astronomia "Galileo Galilei", University of Padua, 35131 Padova, Italy}
\affiliation{Istituto Nazionale di Fisica Nucleare (INFN), 35131 Padova, Italy}
\affiliation{Padua Quantum Technologies Research Center, University of Padua, 35131 Padova, Italy}

\begin{abstract}
We present a systematic study of Tensor Network (TN) models — Matrix Product States (MPS) and Tree Tensor Networks (TTN) — for real-time jet tagging in high-energy physics, with a focus on low-latency deployment on Field Programmable Gate Arrays (FPGAs). Motivated by the strict requirements of the HL-LHC Level-1 trigger system, we explore TNs as compact and interpretable alternatives to deep neural networks. Using low-level jet constituent features, our models achieve competitive performance compared to state-of-the-art deep learning classifiers. We investigate post-training quantization to enable hardware-efficient implementations without degrading classification performance or latency. The best-performing models are synthesized to estimate FPGA resource usage, latency, and memory occupancy, demonstrating sub-microsecond latency and supporting the feasibility of online deployment in real-time trigger systems. Overall, this study highlights the potential of TN-based models for fast and resource-efficient inference in low-latency environments.
\end{abstract}

\maketitle

\section{Introduction}
At the Large Hadron Collider (LHC), proton-proton collisions occur at a rate of 40 MHz, generating raw detector readout rates far exceeding what can be stored or processed offline~\cite{evans_lhc_2008}. To reduce this overwhelming data rate to a level manageable for offline storage and analysis, the ATLAS and CMS general-purpose detectors are equipped with a hierarchical trigger system~\cite{oliveira_damazio_atlas_2025,khachatryan_cms_2017}. This consists of a two-level trigger system that selects events in real time, exploiting different information gathered from the detectors. The upcoming High-Luminosity upgrade of the LHC (HL-LHC) will further increase the experimental complexity and data-processing demands of LHC operations, resulting in even higher data rates and more complex collision topologies~\cite{apollinari_high_2015,tomei_cms_2025}. This scenario will impose groundbreaking demands on the real-time data processing capabilities of trigger systems, particularly the first hardware-based Level-1 (L1) trigger~\cite{Aberle:2749422}. Operating under a latency budget of $12.5~\mu\text{s}$~\cite{L1-HL-upgrade, tomei_cms_2025} and without access to full event information, the L1 trigger must identify interesting physics signatures using hardware-based processing, permanently discarding the rest of the events. This type of environment imposes strict constraints on algorithm design, requiring fixed-latency and pipelined implementations tailored specifically to hardware architectures, like Field Programmable Gate Arrays (FPGAs).

Among the physics objects that must be reliably reconstructed under such conditions are jets, which appear as collimated particle sprays in an LHC detector. Their reconstruction relies on combining particle-tracking data with energy deposits measured in the calorimeters across multiple sub-detectors. Identifying whether a jet originates from a light quark, a gluon, or the decay of a heavy particle such as $W$, $Z$, Higgs boson, or top quark, is known as \textit{jet tagging}~\cite{cagnotta_machine_2022}. This task plays a central role in high-energy physics, as accurate jet classification improves background rejection, enhances the sensitivity of searches for new physics, and improves precision measurements of Standard Model processes~\cite{LARKOSKI20201}. Traditionally, jet tagging has been performed in offline reconstruction and analysis, where detailed event information and powerful machine learning (ML) techniques have been applied to improve jet classification performance without real-time constraints~\cite{PhysRevD.101.056019, app122010574, qu2024particletransformerjettagging, Mondal2024}. With the upcoming HL-LHC upgrade, it has become increasingly important to perform sophisticated jet tagging already at Level-1, where algorithms must be implemented on FPGA hardware.

The strict constraints of the L1 trigger system require fast, interpretable, and resource-efficient algorithms capable of running directly on an FPGA. In this setting, various ML models, ranging from graph-based architectures~\cite{Iiyama} to autoencoders~\cite{Govorkova_2022} and transformer architectures~\cite{laatu, boggia_review_2025}, have been explored for tasks such as particle and jet classification~\cite{duarte_fast_2018, jiang_ultra_2024, que2025jedilinearfastefficientgraph}, event reconstruction~\cite{particleflow2025}, and real-time anomaly detection~\cite{Govorkova_2022, Zipper_2024, gandrakota2024realtimeanomalydetectionl1}. To make these models suitable for real-time inference directly on the FPGA hardware, several model compression techniques, such as quantization and pruning, have been explored to ensure deterministic and low-latency inference suitable for deployment in the HL-LHC trigger environment. In the case of jet tagging, several models achieve state-of-the-art performance in the L1 ultra-low-latency setting. These include JEDI-linear networks~\cite{que2025jedilinearfastefficientgraph}, FPGA-optimized MLP-Mixers~\cite{sun2025fastjettaggingmlpmixers} and more recently transformer-based model~\cite{laatu}. Earlier efforts introduced permutation-invariant neural networks~\cite{Odagiu_2024} and JEDI-net architectures~\cite{Moreno_2020, 10.1145/3640464}, which effectively process sparse particle-level inputs.

Although deep neural networks can be optimized for fast inference, their complex and highly nonlinear internal structure often provides limited transparency and makes it challenging to understand their internal decision-making process to solve a specific task. This motivates the exploration of alternative algorithms inspired by quantum physics, in particular, Tensor Network (TN) models~\cite{simoneintrotensor, stoudenmire_2016}. The structured and factorized representation of TNs offers improved transparency and interpretability compared to conventional deep neural networks. Additionally, TNs offer compact parameterization and leverage linear operations, which makes them suitable for developing smarter low-latency models deployable in resource-constrained environments such as next-generation trigger systems.

This work presents a systematic study of two TN architectures: Matrix Product States (MPS)~\cite{perezgarcia2007matrixproductstaterepresentations,cirac_matrix_2021} and Tree Tensor Networks (TTN)~\cite{shi2006ttn,PhysRevB.80.235127ttn,hikihara_automatic_2023,Felser2021}, developed for the task of jet substructure classification using information of jet particle constituent features. For both approaches, we demonstrate classification performance and evaluate resource utilization and latency under L1 trigger constraints.

\section{Methodology}
\begin{figure*}[ht]
    \centering
    \includegraphics[width=0.8\textwidth]{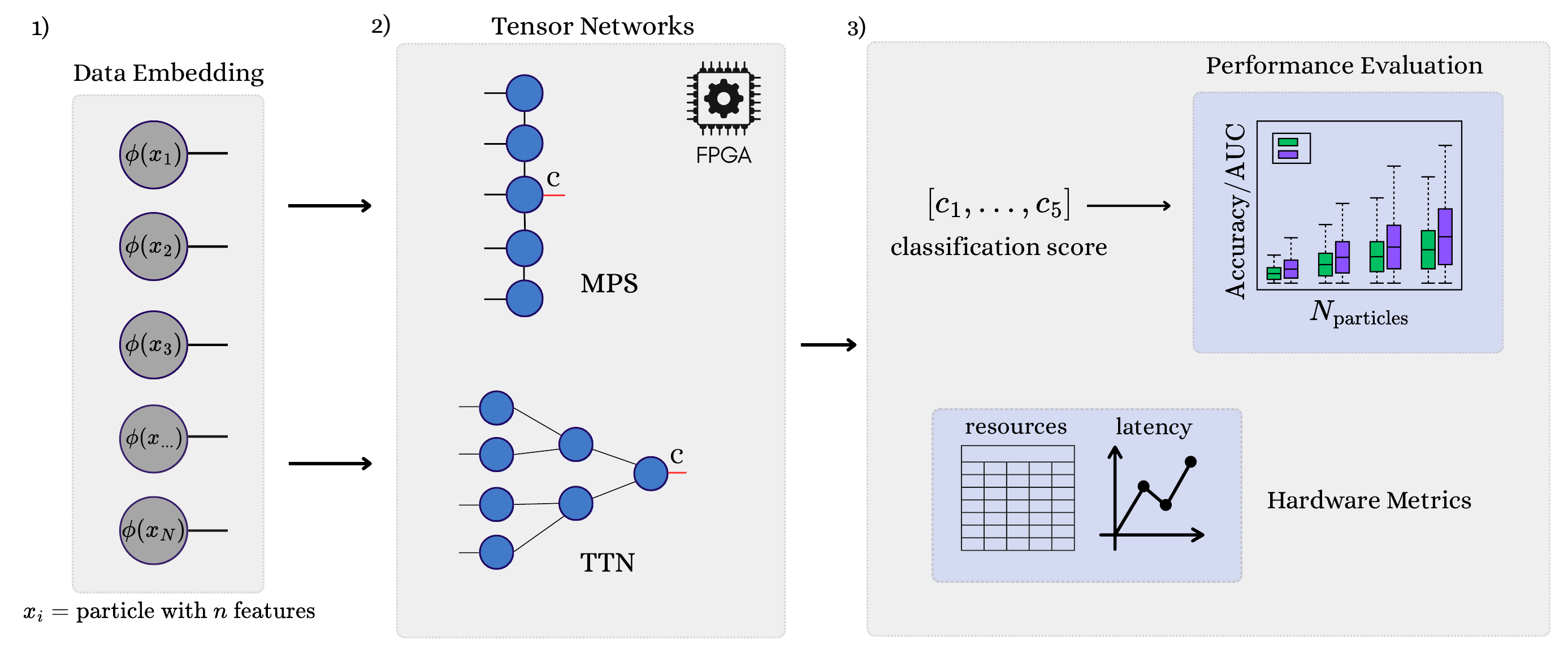}
    \caption{Pipeline of the systematic study consisting of: (1) data embedding procedure embedding each particle $x_i$ ($i \in (1, N)$) with feature map $\phi$ into a product of embedded vectors, (2) Tensor Network model(s) implemented and run on FPGA, (3) performance evaluation in terms of classification accuracy, the area under the receiver operating characteristic curve and relevant hardware metrics. $N$ is the number of particles in a jet. $C$ is an output vector that is used to obtain class probabilities.}
    \label{fig:pipeline}
\end{figure*}
The overall pipeline used in this study to deploy tensor network models on hardware to perform a classification task on a high-energy physics dataset is illustrated in Fig.~\ref{fig:pipeline}.

First, classical data are embedded into the product of tensors using a feature map motivated by quantum information techniques. Next, two TN architectures are designed for the classification task and optimized on the given dataset. Finally, the performance of each TN is systematically evaluated on an FPGA.

\subsection*{Data Representation and Embedding}\label{ssec:embedding}
We use the public \texttt{hls4ml} jet dataset, available on \texttt{Zenodo}~\cite{pierini_2020_3602260}, which provides low-level particle-based representations of jets. Each jet is described by 16 kinematic properties of its particle constituents, enabling the training of TN models directly on the information available at the L1 trigger level. For each jet, the data is represented as a matrix $N \times n$, where $N$ is the number of particles, and $n$ is the number of features per particle. \\
In this work, the first $N$ particles of high-$p_T$ are selected, with $N \in \{8, 16, 32\}$. For each particle, we choose the three most significant features among the $n$ available, namely their transverse momentum ($p_T$), the relative particle energy with respect to the total jet energy ($E_{rel}$), and the distance from the jet center ($\Delta R$). This choice is guided by the results of previous works~\cite{Odagiu_2024, sun2025fastjettaggingmlpmixers}, which highlight the importance of considering both energy- and shape-related features, and the need to reduce the model's computational burden.
This way, one tensor represents one particle, such that it expands the $i^{\text{th}}$ vector of particle features $\vec{x_i}=(p_T, E_{rel}, \Delta R)$ into a polynomial feature map $\phi(\vec{x_i})$ up to degree two:
\begin{equation}
\phi(\vec{x_i}) = \frac{1}{C_i}\left[
1,\ 
p_T,\ E_{rel},\ \Delta R,\ 
p_T^2, \ E_{rel}^2, \ \Delta R^2
\right]~~,
\label{eq:final_embedding}
\end{equation}
where $C$ is a normalization factor.
The final embedding is formed as a tensor product of all individual particle embeddings:
\begin{equation}
    \Phi(\mathbf{x}) = \bigotimes_{i=1}^N \phi(\vec{x_i})
\end{equation}
where each vector in the product has dimension $d=7$. This is the same representation used for a separable state of a quantum many-body system, where each vector in the product corresponds to one \textit{site} with seven levels; from an ML perspective, each site is an input to the model.

The embedding choice adopted in this work is motivated by an analysis of how information is distributed across a family of models trained with different embeddings. The analysis relies on correlations between model inputs, since in tensor networks long-range correlations require larger bond dimensions to be represented faithfully, and therefore a greater number of parameters~\cite{area_laws_RevModPhys.82.277}. As a correlation measure, this study employs the Quantum Mutual Information (QMI), which can be viewed as the quantum-mechanical analogue of Shannon mutual information. Importantly, QMI is evaluated directly on a trained instance of the model, providing insight into how features are internally correlated by the model architecture. As a result, it characterizes correlations that are distinct from the classical correlations present in the input dataset.\\

Unlike in Eq.~\ref{eq:final_embedding}, we first select only $p_T$ and $\Delta R$ for each particle, with each of these features being mapped onto a single site: 
\begin{align}
    \phi_{2i}(\vec{x_i}) &= \frac{1}{C_{2i}}\left[
    1,\ p_T,\ p_T^2 \right] \nonumber \\
    \phi_{2i+1}(\vec{x_i}) &= \frac{1}{C_{2i+1}}\left[
    1,\ \Delta R,\ \Delta R^2 \right].
\end{align}
The information of a single particle is stored in two adjacent sites, thereby doubling the model inputs.
\begin{figure*}[ht]
    \centering
    \subfigure[]{
        \includegraphics[height=8.2cm,trim={0cm 0cm 3.5cm 0cm},clip]{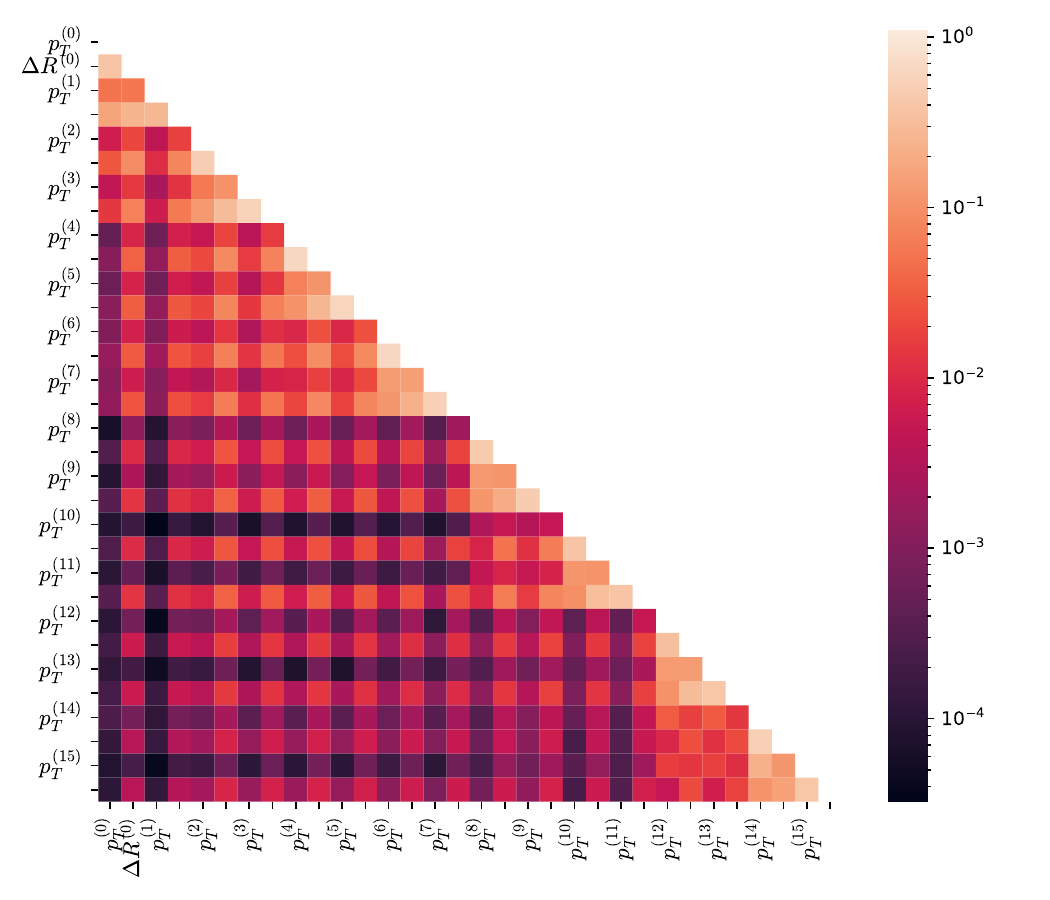}
        \label{fig:qmi_poly}
    }
    \subfigure[]{
        \includegraphics[height=8.2cm,trim={0.25cm 0cm 0cm 0cm},clip]{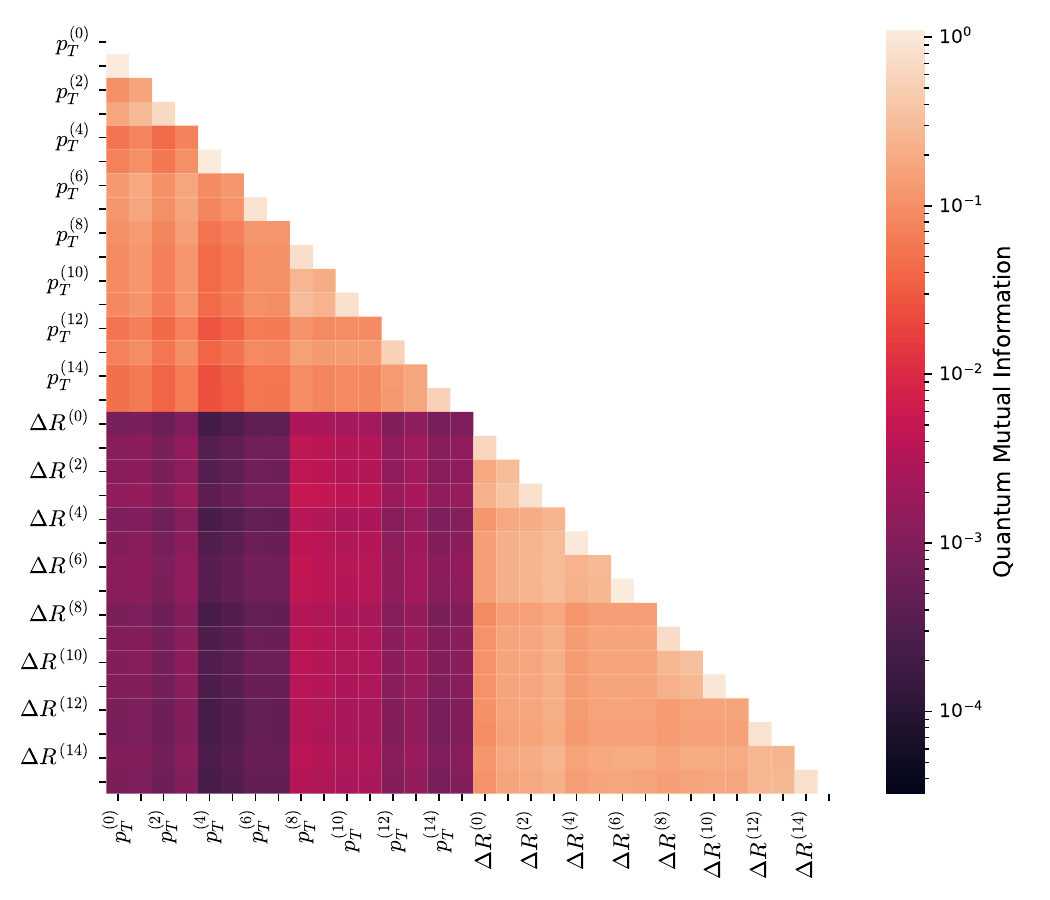}
        \label{fig:qmi_interact}
    }
    \label{fig:qmi}
    \caption{Quantum mutual information between pairs of inputs of a TN model. (a) TN model trained on $N=16$ constituents per jet, embedding $p_T$ and $\Delta R$ of each particle in adjacent sites. (b) TN model trained on the same dataset, where features are permuted so that $p_T$ of all particles come first, followed by $\Delta R$.}
\end{figure*}
In this setting, the QMI exhibits the chessboard-like pattern illustrated in Figure~\ref{fig:qmi_poly}, where long-range correlations are non-negligible. Consequently, we permute the features to bring highly correlated ones close together. Practically, the $p_T$ of each particle is embedded in the first half of the system, and the $\Delta R$ in the second half. The model trained on this particular permutation is characterized by the QMI depicted in Figure~\ref{fig:qmi_interact}. In this case, long-range correlations are diminished, and consequently, a reduced bond dimension is sufficient to maintain the same degree of expressive power.\\
The pattern in Figure~\ref{fig:qmi_interact} suggests that the two halves of the features are treated almost independently by the model. Therefore, the final embedding is designed to stack the features of one particle on one site, without any mixed terms. Rather than solving a problem in an exponentially large space of order $\mathcal{O}(d^{nN})$, the final embedding reduces it to $\mathcal{O}((nd)^N)$, where $n$ is the number of selected features, and $d$ the dimension of the per-feature embedding (in the case of a polynomial embedding, it is the maximum degree of the monomials). The idea of a per-particle embedding was already proposed in Ref.~\cite{bal_1_2025}, while similar results on feature importance were presented in Ref.~\cite{sun2025fastjettaggingmlpmixers}.\\
This discussion highlights the interpretable nature of TNs as a method to gain knowledge about the model's information processing mechanisms and employ it to reduce the number of parameters in the end.

\subsection*{Tensor Network Architectures}
In this study, we explore two different TN architectures: the first being a linear chain of tensors (MPS)~\cite{perezgarcia2007matrixproductstaterepresentations, puljak2025tn4mltensornetworktraining}, and the second organized in a hierarchical binary-tree structure (TTN)~\cite{shi2006ttn}. Here, we describe the specific design of these architectures as used in this study, while a general overview of the models is provided in Appendix~\ref{appendix:tn_models}.\\

\begin{figure}[t]
    \centering
    \includegraphics[width=0.65\linewidth]{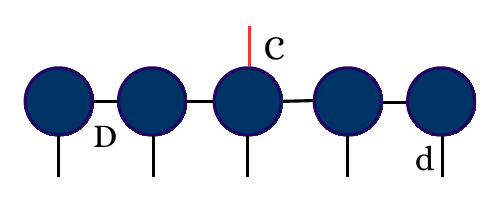}
    \caption{\textit{Matrix Product State} (MPS) structure with five tensors connected via bond indices of dimension $D$, where each tensor has a physical dimension $d$. The central tensor has an additional output leg corresponding to the class index $C$.}
    \label{fig:mps_model}
\end{figure}

The \textbf{Matrix Product State} model used for classification consists of $N$ tensors connected in a chain-like structure, where $N$ is the number of particles per jet (illustrated in Fig.~\ref{fig:mps_model}). Each tensor has a physical dimension $d$ corresponding to the size of the local feature space, and a bond dimension $D$, which controls the model's expressivity and the amount of entanglement it can capture~\cite{cirac_matrix_2021}.
The four-index tensor placed within the chain is responsible for storing the output vector, with the dimension of the external index being the number of classes considered. Tensors are initialized with normally distributed random values with a standard deviation of $\sigma=10^{-2}$. Classification is performed by contracting the embedded input state with the MPS model to obtain a vector of size equal to the number of classes, which is then passed through a softmax function to produce class probabilities. The contraction scheme, illustrated in Fig.~\ref{fig:inference}, is implemented on hardware to maximize the parallelization of operations and minimize inference latency. To improve numerical stability and reduce computational costs, MPS tensors are brought into canonical form, with the canonical center placed at the tensor carrying the class index~\cite{perezgarcia2007matrixproductstaterepresentations,shi2006ttn}. In this form, tensors away from the center are orthonormal, allowing contractions of the remaining tensors outside the canonical center with their hermitian conjugate to reduce to identities.

\begin{figure}[t]
    \centering
    \includegraphics[width=0.7\linewidth, trim={0cm 0cm 0cm 0.6cm}]{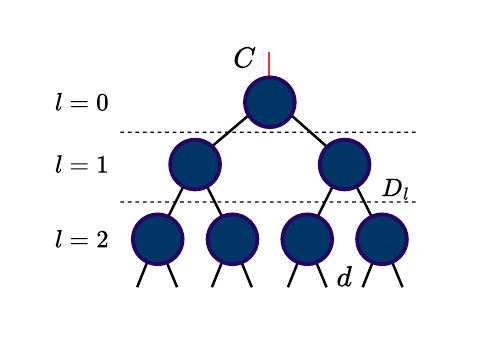}
    \caption{\textit{Tree Tensor Network} (TTN) approximating a $8$-order tensor with three layers $l$, bond dimension $D_l$ for each layer and output index $C$ with dimension corresponding to the number of classes.}
    \label{fig:ttn_model}
\end{figure}

The \textbf{Tree Tensor Network} architecture follows a binary tree structure, where information from the embedded inputs is successively merged through contractions with rank-3 tensors of variable bond dimensions. Tensors are organized in $L=\log_2{N}$ layers, where the ones in the lowest layer have two physical indices of dimension $d$ and one virtual index, while the tensors in the higher layers have three virtual indices with increasing bond dimension $D_l$, up to a maximum value of $\chi$, $D_l=\min(d^{2^{L-l-1}},\chi)$ (see model in Fig.~\ref{fig:ttn_model}).
All tensors are initialized following the procedure described in Ref.~\cite{Stoudenmire_2018}. Conceptually, the layers of a TTN can be interpreted as the progressive application of a kernel function that hierarchically compresses information from the exponentially large Hilbert space into a lower-dimensional representation. A top tensor is included to introduce an external output dimension, which stores classification results and supports multiclass tasks. During inference, the complete contraction between the model and an input sample produces a vector of overlaps, whose squared elements are proportional to the class probabilities.

\subsection*{Training and Evaluation Setup}
Input features per particle are scaled to $[5, 95]\%$ interquantile range to reduce sensitivity to outliers, following Ref.~\cite{Odagiu_2024}. When necessary, jets are zero-padded up to $N$ constituents to ensure a fixed input size. Both models are trained on 620,000 samples and evaluated on an independent test set of 260,000 samples. Training is performed using mini-batch gradient descent and the Adam optimizer. The \textit{cross-entropy} loss function is used for the MPS model, while the TTN model is optimized with a mean squared error loss. This choice was determined empirically based on model stability and convergence behavior. 
Two different software frameworks are used for model implementation:
\begin{itemize}
    \item [-] MPS $\rightarrow$ implemented using \texttt{tn4ml} Python library for building ML pipelines for TNs~\cite{puljak2025tn4mltensornetworktraining}.
    \item [-] TTN $\rightarrow$ implemented using \texttt{qtealeaves} Python library for TN methods, originally developed to simulate many-body quantum systems, with an extension to ML applications~\cite{ballarin_2024_10498929}.\\
\end{itemize}

For the evaluation setup, both models are quantized post-training with a fixed integer bit width of two and a varying fractional bit width. This procedure enables a systematic study of the trade-off between numerical precision, model size, and inference latency. We quantize the models and perform quantized contractions to emulate the arithmetic performed in hardware, ensuring that the reported performance reflects realistic deployment conditions.

To assess robustness under reduced precision, each quantized configuration is benchmarked in terms of classification accuracy. Eventually, we choose the optimal fractional bit width to quantize final models, which are further ported to the FPGA, where we analyze latency and resource consumption performance parameters.

\subsection*{Firmware Implementation on FPGA}

\begin{figure*}[ht]
    \centering
    \includegraphics[width=0.8\linewidth]{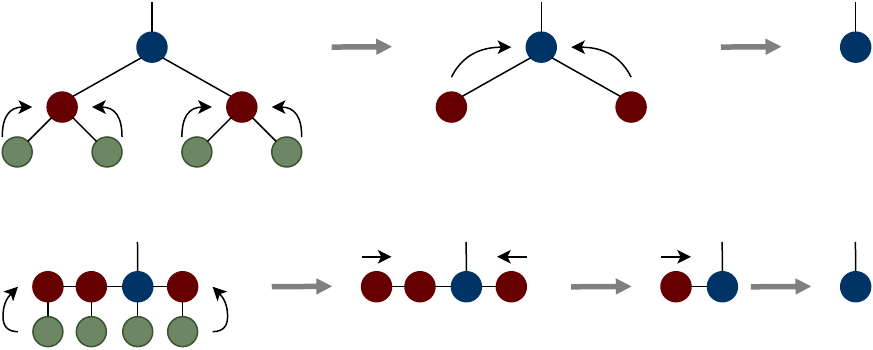}
    \caption{Examples of TTN (top) and MPS (bottom) inference algorithms, considering $N=4$ particles. Black arrows correspond to parallel contractions, while gray arrows link different sequential steps.}
    \label{fig:inference}
\end{figure*}

To perform inference on the FPGA, we implement two distinct algorithms corresponding to the MPS and TTN models. For the MPS, we first contract all the embedded input features with the corresponding MPS tensors. We then perform two parallel chains of contractions: one starting from the left boundary of the MPS and propagating toward the label tensor, and another from the right boundary and converging to the same tensor (see the bottom panel of Fig.\ref{fig:inference}. For the TTN, inference is described as a sequence of chained tensor contractions across the layers of the tree: starting from the lowest layer containing the embedded input feature tensor and ending at the topmost tensor, which encodes the output classification probabilities (see the top panel of Fig.\ref{fig:inference}.

The firmware for TTN inference was developed in VHSIC Hardware Description Language (VHDL) using Vivado~\cite{vivado2024}, and its implementation is described in detail in Ref.~\cite{Borella_2025}. While this low-level approach provides fine-grained control over FPGA resource usage and operation timing, it also requires specialized expertise, which can present a barrier to entry for machine learning practitioners, who are typically more familiar with high-level programming environments. To address this, the MPS firmware was implemented in High-Level Synthesis (HLS) language with Vitis HLS~\cite{amd_vitis_hls_2024_1}, favoring a more accessible and widely usable development framework over low-level hardware design. Both implementations were simulated, validated, and eventually synthesized targeting an AMD-Xilinx XCVU13P FPGA~\cite{noauthor_amd_nodate}.

To test only the behavior and occupancy of the inference algorithms, both MPS and TTN firmware assume that the data pre- and post-processing are handled externally to the FPGA. The input embedding is therefore not implemented in hardware, and each feature is provided to the FPGA already as a $d$-dimensional vector. Similarly, no \texttt{argmax} function is applied to the output to extract the label. The final classification decision is performed off-chip, based on the probability vector produced by the FPGA inference.

\section{Results}
First, we perform hyperparameter optimization for both models in software using floating-point precision until the desired classification performance is achieved. Next, we compare quantization strategies to reduce model size, and the best-performing quantized models are subsequently deployed on the FPGA to evaluate hardware-specific metrics.

\subsection*{Full precision Model Performance}

To evaluate the performance of models in full-precision (\textit{i.e.} 32-bit floating-point arithmetic), Table~\ref{tab:float-performance} presents the results for varying numbers of jet constituents $N$ by reporting classification accuracy and AUC scores for each class. The results reported are produced from MPSs with bond dimension $D=10$ and TTNs with maximum bond dimension $\chi=10$. Compared with permutation-invariant deep learning models in Ref.~\cite{Odagiu_2024}, our TN approaches achieve competitive performance. Furthermore, the performance obtained is close to that of the state-of-the-art transformer and MLP-Mixer models in Ref.~\cite{sun2025fastjettaggingmlpmixers,laatu}. This highlights the competitiveness of TNs compared to conventional neural network architectures and their potential to serve as effective replacements in resource-constrained hardware environments. The inherent linearity and interpretability of TNs facilitate the process of model compression, thereby enhancing their efficacy (see QMI analysis in Sec.~\ref{ssec:embedding}).

\begin{table}[ht]
\centering
\begin{adjustbox}{width=\linewidth}
\begin{tabular}{|c|c|c|c|c|c|c|c|c|}
\hline
\multirow{2}{*}{Model} & \multirow{2}{*}{$N$} & \multirow{2}{*}{$\#$params} & \multirow{2}{*}{Acc(\%)} & \multicolumn{5}{c|}{AUC(\%)} \\
\cline{5-9}
 &  &  &  & $g$ & $q$ & $W$ & $Z$ & $t$ \\
\hline
MPS & \multirow{2}{*}{8} & 6678 & 66.1 & 89.1 & 86.3 & 87.1 & 85.4 & 91.2 \\
TTN &  & 4460 & 65.3 & 89.3 & 87.3 & 87.0 & 84.3 & 92.4 \\
\hline
MPS & \multirow{2}{*}{16} & 12278 & 72.0 & 90.4 & 87.6 & 91.6 & 89.0 & 92.9 \\
TTN &  & 10420 & 72.5 & 91.3 & 89.5 & 91.7 & 89.7 & 93.9 \\
\hline
MPS & \multirow{2}{*}{32} & 23478 & 74.8 & 90.8 & 88.2 & 93.7 & 91.5 & 93.5 \\
TTN &  & 22340 & 77.1 & 92.6 & 90.4 & 94.3 & 93.1 & 94.7 \\
\hline 
\end{tabular}
\end{adjustbox}
\caption{\textit{Full-precision models.} Test accuracy and AUC scores per class for different numbers of particles $N$ per jet. A 3-fold cross-validation procedure was performed. For all models, the errors are smaller or equal to $0.1\%$ for both the accuracy and AUC scores; thus, all are discarded from the table.}
\label{tab:float-performance}
\end{table}

\subsection*{Quantization Study}
Quantization of the tensor weights further reduces the overall model memory footprint by representing parameters in a fixed-point format. We compare the effect of post-training quantization (PTQ) strategy across different bit widths for both the TTN and MPS models. The integer part of the fixed-point representation is set to two bits, while the number of fractional bits (FB) is varied. 

\begin{figure}[ht]
    \centering
    \includegraphics[width=0.9\linewidth]{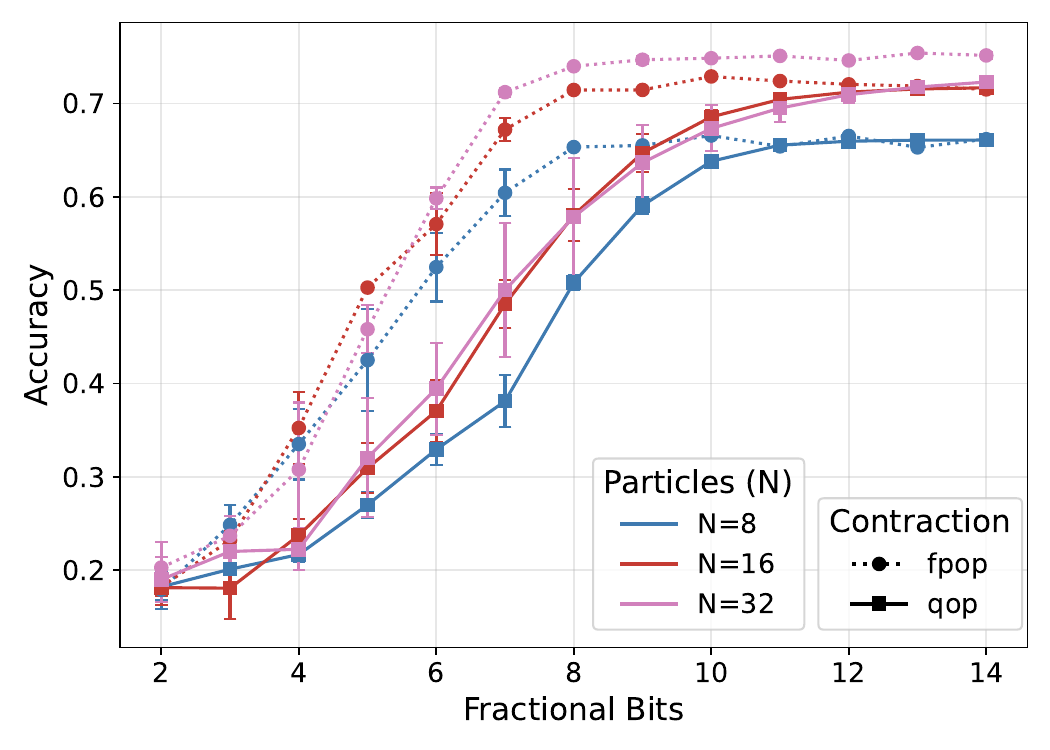}
    \caption{Dependence of classification accuracy on fractional length of the bit width used for post-training quantization (PTQ) of the MPS model with $N=\{8, 16, 32\}$ particles per jet, and bond dimensions $D = 10$. The solid line represents tensor network contractions performed in full precision (fpop) when computing the output vector, while the dotted line corresponds to quantized operations (qop) executed with specified fractional bit width.}
    \label{fig:ptq_mps}
\end{figure}

Figure~\ref{fig:ptq_mps} shows the classification accuracy obtained from PTQ on the MPS model for different numbers of particles per jet ($N$) as a function of the fractional bit width. Additionally, we evaluated the quantized model’s performance with both quantized (solid) and floating-point (dotted) contractions to quantify the impact of reduced-precision arithmetic on the inference and to characterize the degradation introduced specifically by the quantized operations. In practice, the quantization is simulated via software by clipping the results of the operations performed in floating-point. While in the quantized contractions, both model weights and operations are clipped, for floating-point contractions, only the weights are clipped. For all tested values of $N$, the model performance begins to degrade noticeably when the fractional precision is reduced below $\text{FB} = 8$.

\begin{figure}[ht]
    \centering
    \includegraphics[width=0.9\linewidth]{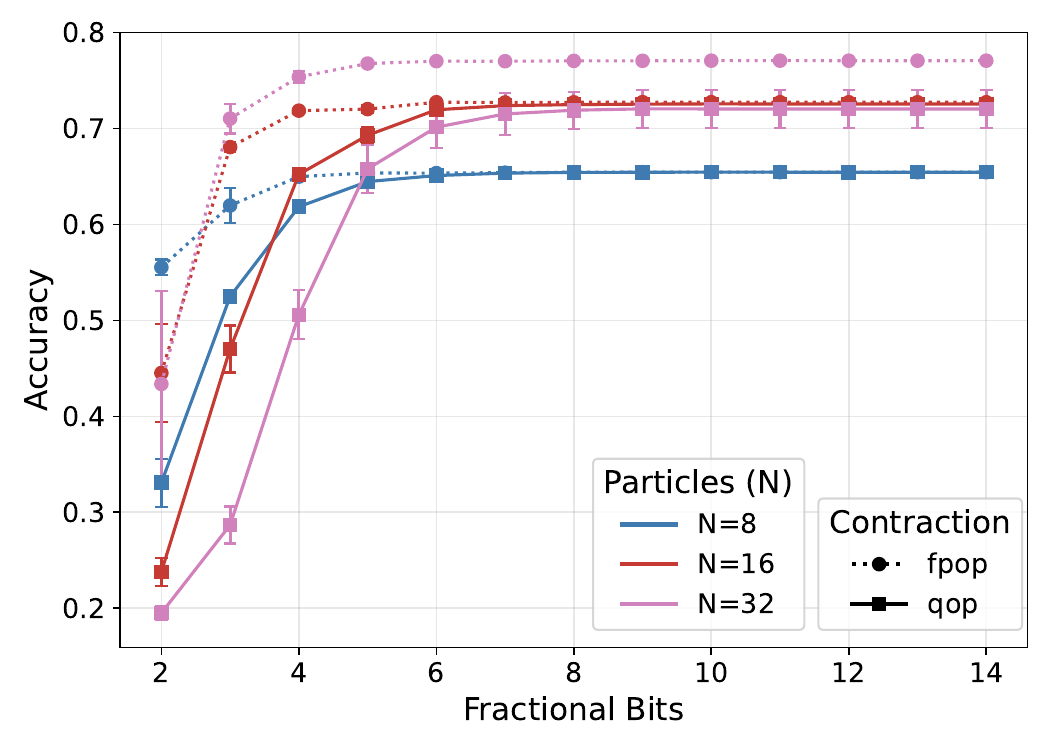}
    \caption{Dependence of classification accuracy on fractional length of the bit width used for PTQ of the TTN model $N=\{8, 16, 32\}$ particles per jet, and maximum bond dimension $\chi = 10$. The solid line represents tensor network contractions performed in full precision (fpop) when computing the output vector, while the dotted line corresponds to quantized operations (qop) executed with specified fractional bit width.}
    \label{fig:quant_ttn}
\end{figure}

Fig.~\ref{fig:quant_ttn} shows the classification accuracy of the TTN models quantized with the PTQ strategy as a function of the number of fractional bits. Quantized contractions (solid line) slightly degrade the performance, but for both types of operations, the classification performance starts to drop more significantly after degrading the fractional bit width below $\text{FB} = 6$.

Overall, the TTN model reaches higher accuracy and reaches its performance plateau with fewer fractional bits, demonstrating stronger tolerance to low-precision arithmetic. On the contrary, the MPS model requires more fractional precision for quantized contractions to align with full-precision results.

\subsection*{Hardware Analysis}
After performing quantization studies separately for MPS and TTN, both models are synthesized for implementation on the XCVU13P device. The resulting occupancy and latency values are reported in Table~\ref{tab:FPGA_resources}. Following the PTQ plots, each model is synthesized both with a maximum fractional precision of 14 bits and with a reduced fractional bit width. In this way, we demonstrate how quantization studies can effectively help in compressing the hardware models without sacrificing classification performance.

The MPS results are obtained from the Vitis HLS synthesis report, which provides an estimate of the final implementation, while the TTN values are derived from the Vivado utilization report generated after synthesizing the VHDL firmware described in Ref.~\cite{Borella_2025}. Since the VHDL implementation allows for complete control over hardware resources instantiation, the TTN occupancy and latency values are fully deterministic and reproducible. Once the tree architecture is fixed (number of particles $N$, number of layers, maximum bond dimension $\chi$, physical dimension $d$), it is possible to derive a precise estimation of the number of Digital Signal Processors (DSP) and the overall inference latency without needing to complete a full synthesis. 

In contrast, the higher-level development environment of Vitis HLS expresses functions and algorithms in C++, where timing behavior is controlled through general-purpose pragmas. These directives allow the user to easily manage the scheduling of operations, their degree of parallelization, and overall resource usage. However, because the translation from high-level algorithmic description in C++ to low-level firmware synthesis is handled by the compiler, resource utilization and latency are difficult to determine a priori. These values must instead be retrieved after synthesis, as different optimization strategies or scheduling decisions may be applied for each model configuration. 

\begin{table*}[ht]
\centering
\begin{tabular}{|c|c|c|c|c|c|c|c|c|c|c|c|}
\hline
N & Model & Params. & FB & Memory [kB] & BRAM [\%] & CARRY8 [\%]& LUT [\%]& FF [\%]& DSPs [\%]& Latency [ns] & $n_{reg}$\\

\hline\hline
\multirow{4}{*}{8} & \multirow{2}{*}{MPS} & \multirow{2}{*}{6678} & 14  & 114 & 0 & - & 1.00 & 4.00 & 52.00 & \multirow{2}{*}{236} & \multirow{2}{*}{3}\\
\cline{4-10}
 & &  & 8 & 78 & 0 & - & 6.00 & 2.00 & 37.00 &  & \\
\cline{2-12}

 & \multirow{2}{*}{TTN} &\multirow{2}{*}{4460} & 14 & 71 & 0 & 4.02 & 4.03 & 2.02 & 39.75 & \multirow{2}{*}{92} & \multirow{2}{*}{1}\\
\cline{4-10}
 &  & & 6 & 35 & 1.38 & 0 & 7.28 & 2.39 & 0.00 &  &  \\
\hline\hline

\multirow{4}{*}{16} & \multirow{2}{*}{MPS} &\multirow{2}{*}{12278}& 14 & 203 & 0 & - & 2.00 & 2.00 & 78.00 & \multirow{2}{*}{432} & \multirow{2}{*}{3}\\
\cline{4-10}
 &  & & 8 & 140 & 0 & - & 8.00 & 3.00 & 58.00 &  & \\
\cline{2-12}
& \multirow{2}{*}{TTN} &\multirow{2}{*}{10420} & 14 & 166 & 0 & 9.41 & 9.44 & 4.71 & 92.41 & \multirow{2}{*}{124} & \multirow{2}{*}{1}\\
\cline{4-10}
 &  & & 6 & 83 & 1.38 & 14.37 & 14.73 & 4.71 & 0.00 &  & \\
\hline\hline

\multirow{4}{*}{32} & \multirow{2}{*}{MPS} &\multirow{2}{*}{23478}& \textit{14} & \textit{383} & \textit{0} & - & \textit{4.00} & \textit{13.00} & \textit{129.00} & \multirow{2}{*}{708} & \multirow{2}{*}{3}\\
\cline{4-10}
 &  & & 8 & 239 & 0 & - & 25.00 & 9.00 & 60.00 &  & \\
\cline{2-12}
& \multirow{2}{*}{TTN} &\multirow{2}{*}{22340} & 14 & 357 & 1.38 & 30.88 & 27.55 & 13.25 & 99.27 & \multirow{2}{*}{156} & \multirow{2}{*}{1}\\
\cline{4-10}
 &  & & 6 & 178 & 1.38 & 29.85 & 29.39 & 9.24 & 0.00 &  & \\
\hline
\end{tabular}
\caption{TTN and MPS models as synthesized on the XCVU13P FPGA with a 250 MHz clock frequency, and quantized to the corresponding fractional bit width (FB) suggested by post-training quantization analysis. The table reports the inference latency and resource utilization (Memory, BRAM, CARRY8, LUT, FF, and DSPs). $n_{reg}$ represents the number of clock cycles needed to perform a multiplication. Non-synthesizable models are reported in italics.}
\label{tab:FPGA_resources}
\end{table*}

In Tab.\ref{tab:FPGA_resources} it is shown how the total resource usage naturally increases with the number of model parameters and with the bit width used to represent them. In particular, all the networks considered in this work manage to fit within the resource limits of the XCVU13P, except for the MPS configuration with $N=32$ and $FB=14$, which exceeds the available number of DSPs. Nonetheless, reducing fractional bit width leads to a significant decrease in the DSP consumption for all models, together with a relatively small increase in the number of Look Up Tables (LUT)~\cite{Borella_2025}. Consequently, the MPS model with $N=32$ can still be deployed on the FPGA using a reduced numerical precision, moving the DSP usage percentage from $129\%$ to $60\%$.

Beyond the intrinsic architectural differences between MPS and TTN, the reported hardware occupancy and latency values also reflect the behavior of the two different hardware synthesis processes. For instance, the TTN models, implemented in VHDL firmware, manage to completely avoid the DSP usage, whereas the HLS-based synthesis of the MPS networks only reduces their DSP deployment.  Moreover, Vitis HLS does not allow explicit control over the number of registers per DSP. Instead, this value is either automatically inferred or set to the maximum during synthesis based on the model's size and structure. While this optimization helps the tool to meet timing constraints, it also scales the inference latency by a factor $n_{reg}$. Additionally, the HLS utilization report does not report estimates for CARRY8 usage, whereas this information is included in the Vivado report for the TTN implementation. 
All synthesized models exhibit an inference latency of $\mathcal{O}(100)$ns and fit within the resource limits of the target FPGA, confirming their suitability for deployment in the Trigger pipeline of HL-LHC general-purpose detectors.

\section{Conclusion and Outlook}
In this work, we establish TN models as a viable and practical option for real-time machine learning inference at the trigger level of HEP experiments. We develop a pipeline for both MPS and TTN classifiers, encompassing all aspects from data embedding to FPGA synthesis. Additionally, these TNs are interpretable by design: the analysis of quantum mutual information provides insights into correlations of the inputs that can be leveraged for model compression and physics-motivated architecture design~\cite{Felser2021}. We demonstrate that TN architectures can meet the constrained environment of the Level-1 Trigger of CMS, both in terms of resource usage and latency. Furthermore, software simulations suggest that these models are robust to numerical quantization.

Motivated by these properties, we further investigate the behaviour of TN models under different hardware implementation strategies. The MPS classifier is synthesized using HLS, thus offering a flexible and programmer-friendly development workflow. In contrast, the TTN is implemented directly in VHDL, providing explicit control over low-level hardware components and ensuring fully deterministic latency and resource usage. Together, these implementations span the spectrum of design choices available for future TN-based trigger algorithms, with one option prioritizing full hardware control and the other facilitating rapid iteration. Guided by the results of post-training quantization simulations, we also synthesize reduced precision models that significantly reduce hardware occupancy while maintaining latency and reducing accuracy slightly.

The best-performing configurations achieve test accuracies of $66.1\%$, $72.5\%$, and $77.1\%$ when utilizing information from 8, 16, and 32 constituents, respectively.
All hardware realizations achieve sub-microsecond inference latency, satisfying the requirements for deployment in real-time trigger systems.

This study provides both a methodology and a hardware-validated proof of concept that supports the adoption of these architectures for real-time inference in low-latency and sensitive environments. Future studies may investigate more advanced numerical quantization techniques~\cite{chang_HGQ2_2024}, or the development of ad hoc post-training quantization methods exploiting the gauge freedom inherent in TN architectures. A quantum-inspired measure of feature importance, combined with quantization methods, has the potential to compress these models further and achieve accurate, efficient, and ultra-fast inference. 

\section{Acknowledgments}
The research leading to these results has received funding from: the Eric \& Wendy Schmidt Fund for Strategic Innovation through the CERN Next Generation Triggers project under grant agreement number SIF-2023-004; the Italian Research Center on High Performance Computing, Big Data and Quantum Computing (ICSC), funded by the European Union through NextGenerationEU (Project No. CN00000013); the European projects EuRyQa (Horizon 2020) and PASQuanS2 (Quantum Technology Flagship); and the Italian Ministry of University and Research (MUR) via the “Quantum Frontiers” project under the Departments of Excellence 2023–2027. Additional support was provided by the German Federal Ministry of Education and Research (BMBF) through the project QRydDemo (funding program “Quantum Technologies – From Basic Research to Market”), the World Class Research Infrastructure – Quantum Computing and Simulation Center (QCSC) of the University of Padova, and the Istituto Nazionale di Fisica Nucleare (INFN) through the iniziativa specifica IS-QUANTUM.
We acknowledge computational resources from CERN and INFN Padova HPC cluster.


\appendix
\section{Tensor Network Models}
\label{appendix:tn_models}
Tensor networks represent a structured factorization of high-order tensors by a graph of lower-order tensors (nodes) linked through contracted indices (edges). As originally developed in condensed matter physics, TNs enable compact representations of high-dimensional data or quantum states in exponentially large Hilbert spaces by leveraging low-rank structure and limited entanglement.

In ML and data-modeling contexts, TNs can often offer compact approximations or parametrizations of a large weight tensor or feature-map tensor. For example, given an embedding $\Phi(x)$ of input features $x$ in a high-dimensional space and a weight tensor $W$, the ML model can be written as $f(x) = W\Phi(x)$ and one could represent $W$ with a TN to reduce parameter count and control correlations~\cite{stoudenmire_2016, puljak2025tn4mltensornetworktraining}. Also, embedding $\Phi$ can be represented with a TN-like structure, depending on the problem.

Although there exist different TN topologies, for ML purposes, loop-less structures are particularly well-studied from model customization to training and evaluation. The most widely used loop-less topologies are the Matrix Product State (MPS)~\cite{perezgarcia2007matrixproductstaterepresentations, puljak2025tn4mltensornetworktraining} and the Tree Tensor Network (TTN)~\cite{shi2006ttn}, which are studied in this work. 

\subsection{Matrix Product State}
An MPS, also known as Tensor Train~\cite{novikov2017exponentialmachines}, is a factorization of a large $N$-order tensor into a chain of at most 3-order tensors. Let $T_{i_1i_2 \cdots i_N}$ be a high-order tensor, where physical index $i_k$ has dimension $d$. The MPS decomposition is written as:
\begin{equation}
\begin{aligned}
T_{i_1 \cdots i_N}
&= \sum_{\{\alpha\}}
A^{[1]}_{i_1,\alpha_1}
A^{[2]}_{\alpha_1,i_2,\alpha_2}
\cdots\\
&\quad
A^{[N-1]}_{\alpha_{N-2},i_{N-1},\alpha_{N-1}}
A^{[N]}_{\alpha_{N-1},i_N}
\end{aligned}
\end{equation}
where each $A^{[k]}$ is at most a 3-order tensor and $\alpha_k$ indices are called \textit{bond} dimensions, each with dimension $D_k$. In simple settings, $D_k = D$. Graphically, an MPS is represented as a chain of nodes (tensors), each having one physical leg (corresponding to index $i_k$) and two bond indices, except at two ends of the chain, which have only one bond index. Often, in ML applications such as classification, the MPS includes an additional index on the middle tensor, as illustrated in Fig.~\ref{fig:mps_model} together with other hyperparameters.

\emph{Space complexity and expressivity} A tensor of dimension $d^N$ can be represented exactly by an MPS if the bond dimension is $D=d^{N//2}$. However, when the bond dimension $D$ is limited, the MPS provides a compressed representation with the number of parameters scaling as $\mathcal{O}(NdD^2)$ instead of $\mathcal{O}(d^N)$. In ML contexts, by controlling the size of $D$, one can tune the tradeoff between model complexity and the amount of correlations captured between neighboring partitions of the chain. If $D=1$, the MPS reduces to a simple product state, while as $D$ grows, the model becomes more expressive.

\subsection{Tree Tensor Network}
In the context of quantum many-body physics, TTN is a factorization of a large $N$-order tensor, representing an $N$-body system, with a different topology from MPS. In more mathematically oriented fields, it is also known as a Hierarchical Tucker decomposition. In general, it can be seen as a directed graph where each node, or tensor, has one and only one parent node and a variable number of children, and each link represents an implicitly contracted index. An additional root tensor can be inserted, with no parents, equipped with the additional index for multiclass classification tasks. Then, a layer can be identified as the set of nodes at the same distance from the root node. In this work, only complete binary trees are considered, implying that all the nodes in the graph are $3$-order tensors, one index being the parent's link, and the other two being the children's. In this setting, $L=\log_2{N}$ layers are needed to decompose an $N$-order tensor. The decomposition can then be written as
\begin{equation}
\begin{split}
    T_{i_1 \cdots i_N} \simeq
    &\prod_{l=0}^{L-2} \prod_{j=0}^{2^l-1} A^{[l,j]}_{\alpha_{[l+1, 2j]},\alpha_{[l+1, 2j+1]},\alpha_{[l-1, j//2]}} \\
    &\prod_{j=0}^{2^{L-1}-1}A^{[L-1,j]}_{i_{2j}, i_{2j+1},\alpha_{[l-1, j//2]}}
\end{split}
\end{equation}

where $l$ is an index running on the layers and $j$ runs on the tensors of the layer. A graphical representation is presented in Fig.~\ref{fig:ttn_model}.

Repeated indices are implicitly contracted, and the non-contracted leg $\alpha_{[-1,0]}$ is the class index. Note that the contraction of this network with the embedded data on the physical links can be executed in parallel on each branch. Furthermore, MPSs can be seen as a special case of TTNs, where each layer is composed of two tensors, with one child index being virtual and the other one being a physical leg.

\emph{Space complexity and expressivity} The dimension of the virtual links, \textit{i.e.} the \textit{bond} dimension, is, considering the children leg of node $[l,j]$, $D_l=\min(d^{2^{L-l-1}},\chi)$. $\chi$, called the maximum bond dimension, is the parameter controlling the complexity of the model, since the number of parameters scales as $\mathcal{O}(N\chi^3)$. When $\chi=d^{N/2}$, the decomposition is exact. When $\chi < d^{N/2}$, we limit the amount of correlations captured between bipartitions of the system.

\bibliography{bib}

\end{document}